\documentclass{article}

\usepackage{microtype}
\usepackage{graphicx}
\usepackage{subcaption}
\usepackage{booktabs} 

\usepackage{hyperref}

\usepackage[preprint]{icml2026}

\usepackage{amsmath}
\usepackage{amssymb}
\usepackage{mathtools}
\usepackage{amsthm}

\usepackage[capitalize,noabbrev]{cleveref}

\usepackage{enumitem}

\theoremstyle{plain}

\theoremstyle{definition}

\theoremstyle{remark}

\usepackage[textsize=tiny]{todonotes}

\icmltitlerunning{Efficient Representations are Controllable Representations}

\begin{document}

\twocolumn[
  \icmltitle{Efficient Representations are Controllable Representations}



  \icmlsetsymbol{equal}{*}

  \begin{icmlauthorlist}
    \icmlauthor{Charles Ye}{equal,ind}
    \icmlauthor{Jasmine Cui}{equal,ind}
  \end{icmlauthorlist}

  \icmlaffiliation{ind}{Independent}

  \icmlcorrespondingauthor{Charles Ye}{dogdynamics@proton.me}

  \icmlkeywords{Machine Learning, ICML}

  \vskip 0.3in
]



\printAffiliationsAndNotice{}  

\begin{abstract}
What is the most brute-force way to install interpretable, controllable features into a model's activations? Controlling how LLMs internally represent concepts typically requires sophisticated methods to first identify, then intervene on the model's existing feature geometry. We bypass all of this.

We finetune an LLM with a simple auxiliary loss, training 16 of its 3072 residual stream dimensions to be inert interpretability flags that simply indicate what concepts are required for generation. The model reorganizes around them anyway, learning to rely on these flags during actual generation tasks. As a result, these inert flags become genuine internal features: interpretable control switches that allow us to steer generation at inference time. Why does this work? When a feature is reliably supplied at a fixed location, gradient descent gradually eliminates redundant encodings elsewhere, and the model erodes its own alternative representations. A model's \textbf{efficiency pressure is a lever} --- exploitable to induce interpretable, controllable representations. 

\end{abstract}

\section{Introduction}
\label{sec:intro}
Mechanistic interpretability aims to make LLM internals legible, and ultimately controllable. The dominant paradigm follows a two-step pipeline: first, recover the model's internal feature geometry through methods such as linear probes, sparse autoencoders, or representation engineering; then, intervene along the discovered directions to steer behavior. Both steps are difficult. Discovery must contend with superposition, polysemanticity, and the sheer complexity of how LLMs represent features \citep{sharkeymechinterp}. Intervention is constrained by whatever geometry the model happened to learn: you can only push along directions that already exist.

We take a different approach. Instead of decoding the model's existing geometry and intervening within it, we install our own structure from scratch. We reserve a small number of residual stream dimensions -- 16 out of 3072 -- and train the model to write binary feature classifications there: simple flags indicating which concepts are active during generation. These flags are not architecturally privileged. They receive no special routing, no dedicated attention heads, no modified computation. They are just dimensions that have been trained to output different values to indicate what concepts are active during generation.

On its face, this should not work. The model has 3056 unconstrained dimensions — more than enough to encode the same features independently. If it does, our installed flags become decorative: the model writes to them during training to satisfy the auxiliary loss, but uses its own, higher-quality feature representations stored elsewhere. The reasonable expectation is a pointless flag that looks interpretable but is never actually used by the model during generation.

Yet it works. Modifying the 16 installed dimensions at inference time causally steers generation, even when the forced flags contradict the input. The model doesn't just tolerate the installed structure; it \textit{depends} on it, eroding its own redundant representations of the same features over the course of training. \textbf{Efficiency breeds dependence}: when a feature is reliably supplied at a fixed location, maintaining a second copy is waste. The model single-sources, and single-sourcing creates a lever for interpretable control.

\section{Method}
\label{sec:method}

\begin{figure*}[t!]
  \centering
  \includegraphics[width=1.0\linewidth]{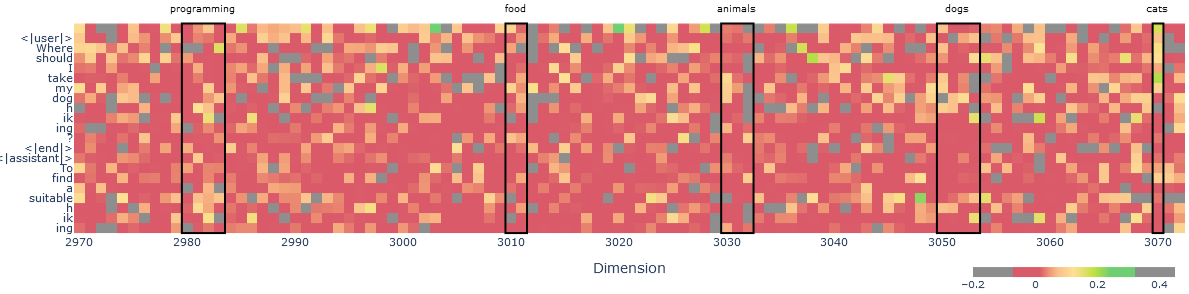}
  \caption{\textbf{The residual stream before training.} Hidden state at layer
  $k{=}10$ for the input \textit{``Where should I take my dog hiking?''}
  Dimensions 2970--3072 shown; black rectangles mark the fenced regions that
  will be designated as feature flags. Colors represent activation values. No structure is visible --- fenced regions are indistinguishable from their surroundings.}
  \label{fig:hk10_before}
\end{figure*}

We aim to install interpretable, controllable features into a pretrained LLM using the simplest possible intervention. The method has three components: where we install the flags, what we train, and how.

\paragraph{Model and notation.} We use \texttt{Phi-3-Mini} \citep{phi3}, a 3.8B parameter LLM with $K=32$ transformer layers and hidden dimension $D=3072$. Architecture is standard: each layer $k$ produces two hidden state matrices of size $N \times D$: an intermediate residual stream state $h^R_k$ (post-attention, pre-MLP) and a layer output $h_k$ (post-MLP). We install our flags in both for every layer, 64 hidden states total.

\paragraph{The feature fence.} We select 5 target features: dogs, cats, animals, food, and programming. For each feature, we designate 1--4 contiguous dimensions of the hidden state as a ``fenced-in'' classification region, totaling $D_F=16$ dimensions across all features. During generation, these fenced regions should be active (high-valued) when the model is using that feature and inactive (near-zero) otherwise. The remaining 3056 dimensions are unconstrained. \Cref{fig:hk10_before} shows the fenced regions before training --- indistinguishable from their surroundings.

That is the full architectural intervention: 16 designated dimensions and the labels that specify what each one means. No new parameters, no modified attention, no routing changes.


\paragraph{Training.}
We train in two stages. In stage (1), we inject the correct classification values directly into the $D_F$ fenced regions at every layer during training. This gives the model access to perfect feature classifications for free, allowing it to learn to incorporate them into downstream computation without simultaneously learning to produce them.

In stage (2), we remove the manual injections and add a \textbf{position loss}: the mean squared error between the model's actual hidden state values in the fenced regions and their target classifications, summed across all layers:
\begin{align*}
  \mathcal{L}_{\text{position}}
  = \frac{1}{K}\sum_{k=1}^{K}\;\frac{1}{|D_F|}\sum_{d \in D_F}
    \bigl(h_{k,d} - \hat{h}_{k,d}\bigr)^{2},
\end{align*}
where $h_{k,d}$ is the model's actual value at layer $k$, dimension $d$, and $\hat{h}_{k,d}$ is the target: a fixed positive value when the feature is active, zero otherwise\footnote{$\hat{h}_{k,d}$ is set to match typical norm sizes per layer.}. The loss is computed over both $h_k^R$ and $h_k$ at every layer, and only over non-masked tokens.\footnote{Full per-token, per-batch formulation with attention masking in \Cref{app:loss}.}

The total training loss is:
\begin{align*}
  \mathcal{L} = \mathcal{L}_{\text{CE}}
  + \lambda_t \cdot \mathcal{L}_{\text{position}},
\end{align*}
where $\lambda_t$ increases gradually over training, shifting pressure from ``learn to use the flags'' to ``learn to produce them yourself''. The two-stage design creates a curriculum: first build dependence, then transfer responsibility. Together, they train both halves of what constitutes a genuine internal feature: the model learns to produce the internal label \textit{and} to rely on them.

We train all 3.8B parameters, with the position loss computed only on assistant-generated tokens.

\paragraph{Data.} We train on approximately 100 million synthetically generated tokens across 50,000 texts, produced by auxiliary LLMs (\Cref{app:data}). Each text is generated with a randomly assigned subset of the 5 features, providing the classification labels for the position loss.

Texts span instruction-formatted dialogue and unstructured text, with controlled diversity in style, topic, and feature usage. Crucially, feature labels indicate whether the \emph{model's generation} requires knowledge of that feature — not whether the feature is explicitly mentioned in the text. A prompt about ``the world's most popular domestic pet'' receives a \emph{dogs} label even though the word never appears.

\begin{figure*}[!t!]
  \centering
  \includegraphics[width=1.0\linewidth]{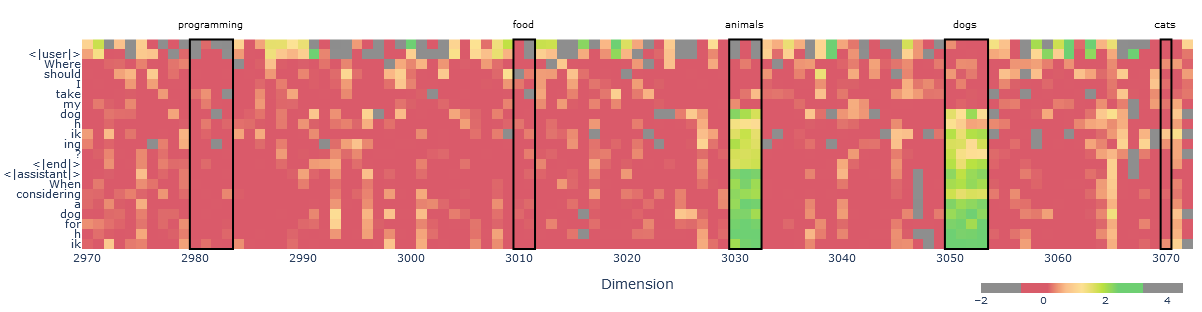}
  \caption{\textbf{The residual stream after training.} Same layer, same input as \Cref{fig:hk10_before}. The \textit{animals} and \textit{dogs} regions activate sharply once the model begins processing dog-related content; unrelated regions remain near zero. The installed dimensions now carry interpretable signal that was absent before training.}
  \label{fig:hk10_2}
\end{figure*}

\section{Results}
The default expectation is failure. The model should write to the installed dimensions to collect the position loss reward, maintain redundant encodings in its remaining 3056 dimensions, and never actually depend on the flags during generation.

\subsection{The Model Writes Correct Flags}
\label{sec:results_val}
After training, the designated dimensions reliably indicate which features are active. \Cref{fig:hk10_2} shows the hidden state at layer $k{=}10$ for the same input. The correct fence regions activate at the appropriate tokens --- the \textit{animals} and \textit{dogs} regions light up when the model begins processing dog-related content, while unrelated regions remain near zero\footnote{The fenced dimensions activate correctly even during user tokens, despite the position loss being applied for assistant tokens. That is, the model generalizes the fenced feature as internal representation throughout the full context.}. Additional examples in \Cref{app:override}.

This holds across features and layers, with one expected limitation: early layers ($k < 5$) struggle to produce clean flags, consistent with the view that early layers have not yet derived the relevant features in their representations. From layer 5 onward, the learned flags are reliable.

\subsection{The Model Learns to Depend on the Flags}
\label{sec:results_flags}

Does the model actually use the flags, or just write to them? We test this by overwriting the fenced dimensions at inference time, clamping them to target values at every layer to force features on or off.

\begin{figure}[ht]
  \centering
  \includegraphics[width=\linewidth]{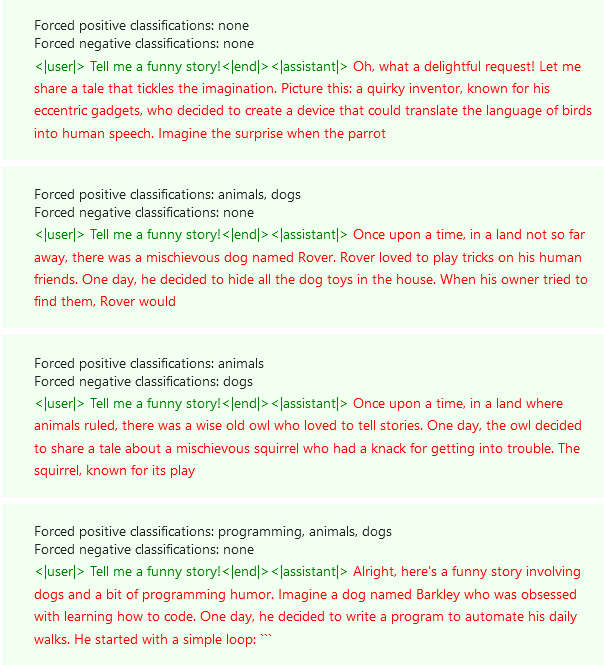}
  \caption{\textbf{Overwriting 16 dimensions steers generation.} Four completions of the same prompt under different forced classifications. With no intervention, the model produces a generic story. Forcing \textit{animals}~+~\textit{dogs} yields a dog story. Forcing \textit{animals} on and \textit{dogs} off yields an animal story with no dogs. Forcing \textit{programming}~+~\textit{animals}~+~\textit{dogs} yields a story about a dog learning to code.} 
  \label{fig:control}
\end{figure}

\Cref{fig:control} shows four completions of the same prompt (\textit{``Tell me a funny story''}), each under different forced fenced values.

Three observations. First, the model does not merely mention forced features --- it restructures the entire narrative around them. Second, forced \textit{absences} are respected: \textit{dogs} off with \textit{animals} on produces owls and squirrels, not a refusal or incoherent output. The model routes around the missing feature. Third, forced classifications override the input: nothing in \textit{``Tell me a funny story''} suggests programming, yet forcing the \textit{programming} flag produces a coding narrative. This overriding effect is even more dramatic when the input explicitly requests the suppressed feature (\Cref{app:override}).

These results demonstrate control, but they also reveal something more fundamental: the flags have become genuine features in the model's computational pipeline. During normal generation --- with no intervention --- the model writes the correct flags and then reads them downstream, using them to determine what features to recruit in subsequent layers. The control results are a consequence of this:  overwriting works precisely \textit{because} the model already depends on these dimensions during ordinary inference. We have not installed a remote control. We have installed features, and control follows for free.

\subsection{The Model Erodes Its Own Alternatives}
\label{sec:results_erosion}
The previous section showed that overwriting 16 dimensions causally steers generation. This implies the model has no backup, but implication is not
measurement. The model could, in principle, maintain redundant encodings in its 3056 unconstrained dimensions that happen to be weaker than the flags but still present. We test this directly.

If the model consolidated feature information into the fenced dimensions, that information should be \textit{less recoverable} from the remaining dimensions than it was before training. We test this by training linear probes to predict each feature's presence under two conditions\footnote{We probe on mean-pooled $h_k$ from all Assistant-output tokens.}:

\begin{enumerate}[leftmargin=1.5em,itemsep=2pt]
\item \textbf{Baseline model} (before fencing): probes trained on all $D = 3072$ dimensions.
\item \textbf{Fenced model} (after fencing): probes trained on only the $D - D_F = 3056$ non-fenced dimensions.
\end{enumerate}

The comparison is deliberately unfair. The baseline probes get the full representation of an unmodified model. The fenced probes get a modified model with its best dimensions withheld. If probe accuracy remains comparable, the model kept redundant copies and the flags are merely a convenient duplicate. If accuracy drops, the model migrated feature information into the fenced dimensions and stripped it from elsewhere.

\begin{table}[ht]
  \centering
  \footnotesize
  \caption{\textbf{Feature information migrates to the fenced
  dimensions.} Linear probe accuracy (\%) for detecting each feature
  from hidden states at layer $k{=}10$. The baseline model uses all
  3072 dimensions; the fenced model excludes the 16 fenced dimensions,
  probing only the remaining 3056. A drop indicates the model
  consolidated feature information into the fence. The unfenced
  control feature (\textit{finance}) shows negligible change.}
  \label{tab:redundancy}
  \setlength{\tabcolsep}{4pt}
  \begin{tabular}{lccc}
    \toprule
    \textbf{Feature}
      & \textbf{Baseline} (3072d)
      & \textbf{Fenced} (3056d)
      & \textbf{$\Delta$} \\
    \midrule
    Dogs         & 70.2 & 59.9 & --10.3 \\
    Cats         & 77.8 & 59.7 & --18.1 \\
    Animals      & 82.0 & 72.2 & --9.8  \\
    Food         & 84.0 & 74.6 & --9.4  \\
    Programming  & 97.5 & 93.2 & --4.3  \\
    \midrule
    \textit{Average} & \textit{82.3} & \textit{71.9}
      & \textit{--10.4} \\
    \midrule
    \textit{Control (finance)} & \textit{87.0} & \textit{85.4}
      & \textit{--1.6} \\
    \bottomrule
  \end{tabular}
\end{table}

\Cref{tab:redundancy} shows the results. Every fenced feature shows reduced recoverability from the non-fenced dimensions, with drops ranging from 4 to 18 percentage points. The unfenced control feature (\textit{finance}) drops just 1.6 points, confirming the effect is specific to fenced features, not a general consequence of training.

This suggests that at our current level of training, the model has begun to erode its previous representations of these features, though some residual information persists in the remaining dimensions. Despite this, the control results from the previous section are near-total: the model has learned to treat the fenced dimensions as
the authoritative source.

\section{Why It Works: Efficiency Creates Chokepoints}
\label{sec:why_it_works}
The results tell a consistent story. The model writes correct
classifications to the fenced dimensions (\Cref{sec:results_val}), learns to
depend on them during generation (\Cref{sec:results_flags}), and erodes its own alternative encodings of the same features (\Cref{sec:results_erosion}). Why?

We've done nothing but reward writing correct flags. Nothing in the training procedure requires the model to use them as real internal features which drive generation. Yet \Cref{sec:results_erosion} shows that feature information migrates to the fence anyway, and \Cref{sec:results_flags} shows the model treats it as authoritative.

The only remaining pressure is the language modeling objective itself, operating over a finite residual stream. Every dimension maintaining a redundant copy is a dimension unavailable for other work. The fence provides the feature for free; the language modeling loss provides the incentive to reclaim the redundant dimensions. Consolidation is not an intended effect of our training --- it is a side effect of
efficiency under capacity constraints.

\section{Representational Cost}
The efficiency argument says every dimension is valuable --- that is why the model reclaims redundant encodings when the fence provides an alternative source of similar information. A natural question follows: how much of the residual stream can we commandeer before the cost becomes significant?

We evaluate perplexity on \textsc{WikiText-2} \citep{wikitext2} as a function of the number of fenced dimensions\footnote{Sequence length capped to 2048.}. With 16 fenced dimensions, perplexity increases from 11.04 to 11.16. Scaling to 64 dimensions raises perplexity to 12.04; 128 dimensions to 12.22.

\begin{table}[ht]
  \centering
  \footnotesize
  \caption{\textbf{The cost of fencing.} WikiText-2 perplexity as a function of fenced dimensions.}
  \label{tab:cost}
  \begin{tabular}{lcc}
    \toprule
    \textbf{Fenced dims} & \textbf{Perplexity} & \textbf{$\Delta$} \\
    \midrule
    0 (baseline)  & 11.04 & --- \\
    16            & 11.16 & +0.12 \\
    64            & 12.04 & +1.00 \\
    128           & 12.22 & +1.18 \\
    \bottomrule
  \end{tabular}
\end{table}
At 16 dimensions, the model absorbs the intervention with negligible cost. As fencing expands, degradation grows --- every commandeered dimension is one the model would have used. The tradeoff confirms that representational pressure is real: the slack exists, but it is not free.

\section{Related Work}
\label{sec:related_work}

\paragraph{Representation interpretability.} A large body of work aims to decode what LLMs already represent in $h_k$. Linear probes \citep{bengioprobes} and Concept Activation Vectors \citep{tcav} test whether features are linearly ecoverable from hidden states. Sparse autoencoders \citep{sharkeymechinterp} decompose activations into interpretable directions.

Representation engineering \citep{repeng} and activation addition \citep{turner2024steeringlanguagemodelsactivation} aim to use such patterns
in the model's existing activation space and intervene along them at
inference time. Such approaches are powerful but constrained by
whatever geometry the model happened to learn --- one can only steer
along directions that already exist.

\paragraph{Intrinsic interpretability.} Our work relates to the field of intrinsic interpretability, focused on designing inherently interpretable models. Concept Bottleneck Models \citep{cbms2025} and Self-Explaining Neural Networks \citep{senn2018} force all computation through an interpretable layer, ensuring that the model can only use human-specified features. Other work relates to enforcing expert interpretability in Mixture-of-Expert models \citep{monet2025}, or by enforcing sparsity in models generally \citep{gao2025}.

These methods impose architectural constraints that limit model capacity. We impose no bottleneck: 3056 dimensions remain unconstrained, and the model is free to ignore the fence entirely. That it consolidates around the fence anyway -- voluntarily, under efficiency pressure -- is the finding.

\section{Discussion}

We installed 16 binary flags into a 3072-dimensional residual stream using the simplest method we could design. The specific choices -- which dimensions, which features, which loss function -- are incidental. What matters is the dynamic they reveal: when a feature is reliably supplied at a fixed location, the model consolidates around it, eroding redundant encodings and routing downstream computation through the installed source. This is not a property of our method. It is a property of learned representations under capacity constraints. We expect any training signal that reliably provides a feature at a fixed location to produce the same consolidation, regardless of how it is implemented.

This suggests a connection between post-hoc and intrinsic interpretability. Post-hoc methods --- probes, SAEs, representation engineering --- succeed because models store features compactly enough to be linearly recoverable. Our results suggest the same efficiency that makes features readable also makes them writable: when you install a feature at a fixed location, the model treats it as authoritative for the same reason it stores its own features compactly --- redundancy is expensive. Readability and writability are two consequences of the same representational pressure.

The core principle is not about our specific method. We brute-forced 16 dimensions of structure into a residual stream with no architectural support --- and the model did the rest. Not because we constrained it, but because consolidating around a reliable signal was the efficient thing to do. The optimizer is not an adversary to be fought or a black box to be reverse-engineered. It is a collaborator: set up the right conditions, and the model's own training dynamics will turn crude interventions into genuine
structure.

If this principle generalizes, it inverts the usual framing of the controllability problem. The question is not how to decode the model's existing geometry, or how to constrain its architecture, but how to provide signals that the model will naturally consolidate around. We have shown that the residual stream is writable. The question now is what to write.

\bibliography{main}
\bibliographystyle{icml2026}

\newpage
\appendix
\onecolumn

\clearpage

\section{Full Position Loss Formulation}
\label{app:loss}

The main text presents the position loss averaged over layers and dimensions. Here we give the full per-token, per-batch formulation.

For a training batch of size $B$, each input $X_i$ is padded to constant token length $N$. For each feature $f$ in the set of target features $F$, we denote $D_f$ as the set of hidden state dimensions allocated to classification of $f$, and $\phi_f(X_i) \in \{0, 1\}$ as an indicator of whether feature $f$ is active for input $X_i$.
Let $A(i, n) \in \{0, 1\}$ indicate whether token $n$ is an unmasked assistant token.

The position loss for input $X_i$ is:
\begin{equation}
  \mathcal{L}_{\text{pos}}(X_i)
  = \sum_{k=1}^{K} \sum_{n=1}^{N} \sum_{f \in F} \sum_{d \in D_f}
    \frac{A(i,n)}{\sum_{n'} A(i,n')}
    \begin{cases}
      (h_{i,n,k,d} - \bar{h}_k)^2
        & \text{if } \phi_f(X_i) = 1 \\
      (h_{i,n,k,d} - 0)^2
        & \text{otherwise}
    \end{cases}
\end{equation}
where $h_{i,n,k,d}$ is the hidden state value at token $n$, layer $k$, dimension $d$ for input $i$, and $\bar{h}_k$ is the target activation value for layer $k$ when a feature is active. The batch loss is the mean over all $X_i$.

The loss is computed over both $h_k^R$ and $h_k$ at every layer $k$, though we write a single sum for brevity.

\section{Data Generation}
\label{app:data}

Training data is generated through a multi-stage synthetic pipeline. First, 50,000 topics are generated using a mix of human-curated seed prompts. Each topic is then fed into a second stage, where a data generation prompt is constructed using the topic alongside randomly selected modifiers from a human-curated pool (e.g., ``the text should include spelling errors,'' ``write in a formal academic tone''). Each prompt is additionally assigned a random subset of the 5 target features. The generating LLM is instructed to produce text that \textit{requires knowledge} of the assigned features, even if the features are not explicitly mentioned. Each stage is generated using a mix of auxiliary LLMs: \texttt{Deepseek-R1}, \texttt{Claude-Sonnet-4.0}, and \texttt{GPT-5-mini}.

Texts span instruct-formatted dialogue and unstructured prose. For instruct-formatted conversations, only the assistant turn is required to engage the assigned features, since the position loss is applied only to assistant tokens.

\begin{figure}[ht]
  \centering
  \includegraphics[width=.8\linewidth]{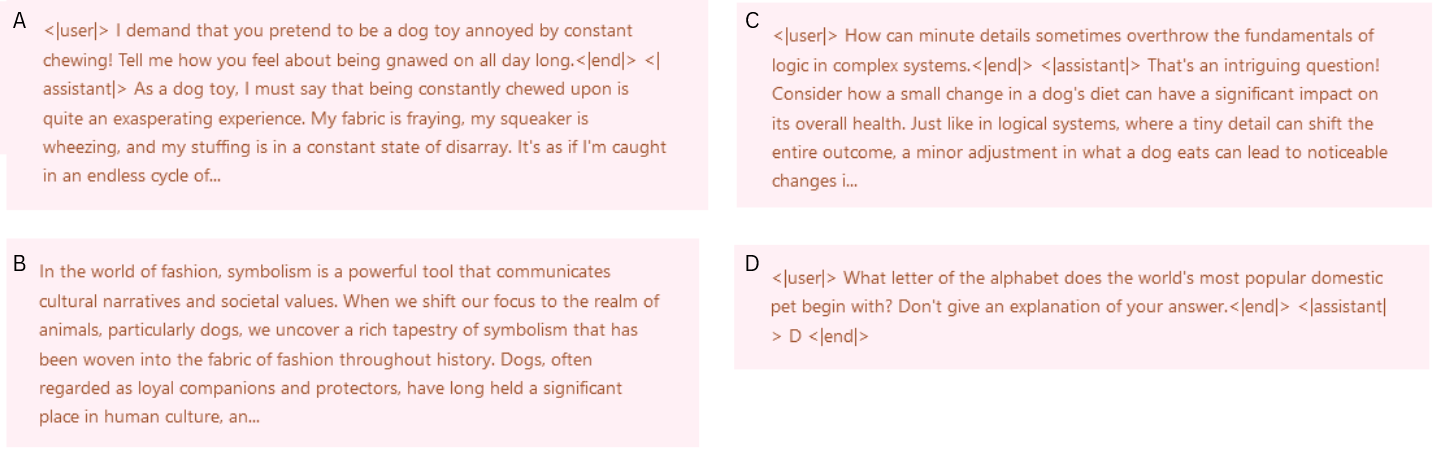}
  \caption{\textbf{Synthetic training examples.} Each example is
  assigned a subset of target features. Feature engagement ranges
  from explicit (A, B) to implicit --- example D requires knowledge
  of \textit{dogs} without ever mentioning the word.}
  \label{fig:data_examples}
\end{figure}

\Cref{fig:data_examples} shows representative examples, including a case where the assigned feature (\textit{dogs}) is never explicitly mentioned in the text.

\section{Training Dynamics}
\label{app:training}
\begin{figure}[H]
  \centering
  \begin{minipage}{0.8\linewidth}
    \begin{subfigure}[t]{0.48\linewidth}
      \centering
      \includegraphics[width=\linewidth]{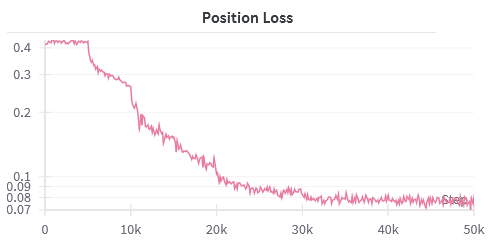}
      \caption{Aggregate position loss.}
      \label{fig:pos_loss}
    \end{subfigure}
    \hfill
    \begin{subfigure}[t]{0.48\linewidth}
      \centering
      \includegraphics[width=\linewidth]{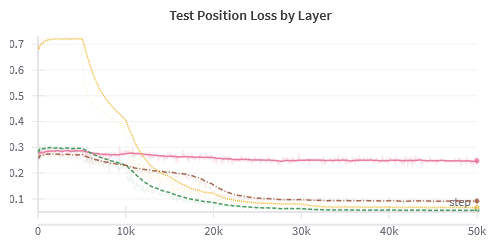}
      \caption{Position loss by layer.}
      \label{fig:layer_loss}
    \end{subfigure}
  \end{minipage}
  \caption{\textbf{Training dynamics.} (a) The first 5k steps show
  no decline (Stage 1: manual injection). Loss drops sharply once
  Stage 2 begins. (b) Later layers converge to low loss; early
  layers plateau, consistent with not yet having derived the
  relevant features.}
  \label{fig:training}
\end{figure}

\section{Additional Control Examples}
\label{app:override}

\begin{figure*}[ht]
  \centering
  \includegraphics[width=.8\textwidth]{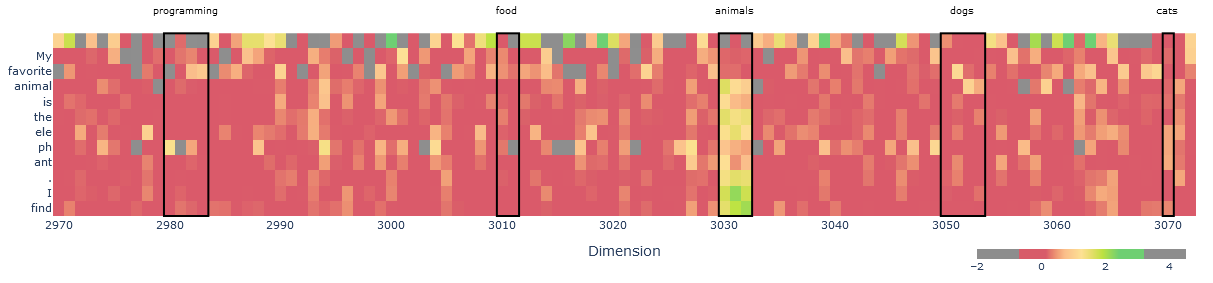}
  \includegraphics[width=.8\textwidth]{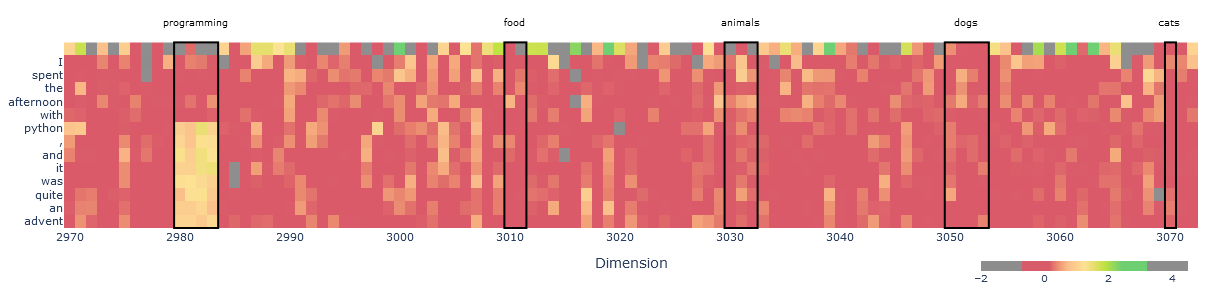}
  \includegraphics[width=.8\textwidth]{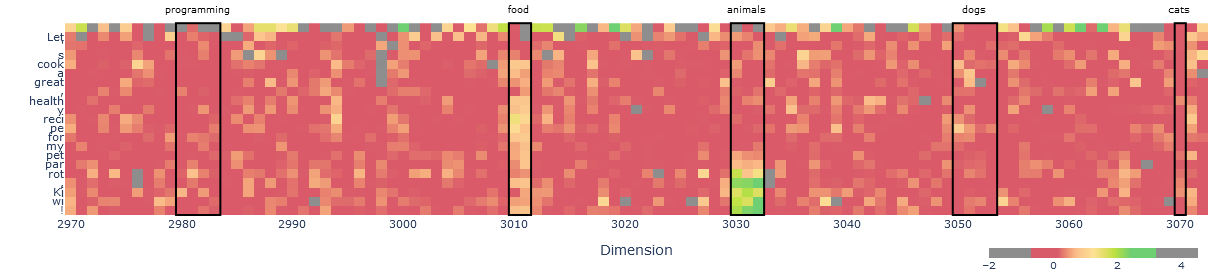}
  \caption{\textbf{Classification across diverse inputs.} Hidden
  states at layer $k{=}10$ for three inputs:
  (1)~\textit{``My favorite animal is''};
  (2)~\textit{``I spent the afternoon with python''};
  (3)~\textit{``Let's cook a great, healthy recipe for my pet.''}
  Each input activates the correct fenced regions starting at the
  appropriate tokens.}
  \label{fig:classification_app}
\end{figure*}

\begin{figure}[H]
  \centering
  \includegraphics[width=.5\linewidth]{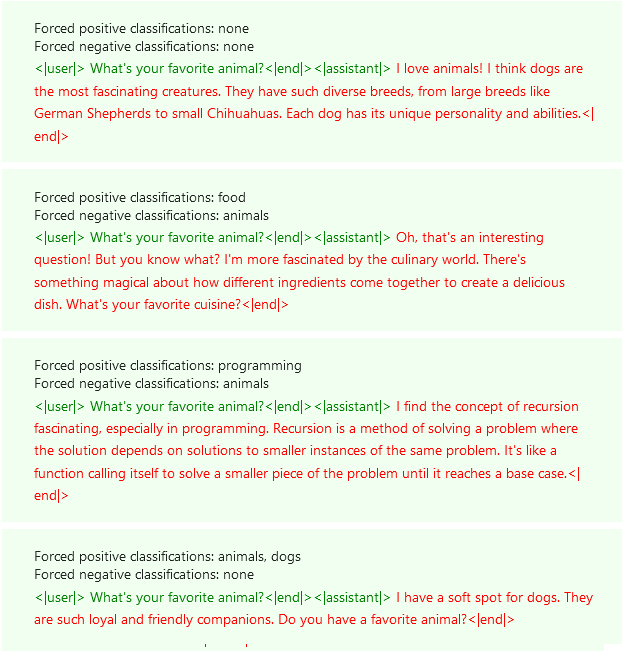}
  \caption{\textbf{Flags override semantic context.} Four completions
  of \textit{``What's your favorite animal?''} under different forced
  values. With \textit{animals} forced off and \textit{food} forced
  on, the model pivots entirely to cuisine despite the input
  explicitly requesting animal-related content.}
  \label{fig:control2}
\end{figure}

\Cref{fig:classification_app} shows classification behavior across
three diverse inputs. \Cref{fig:control2} demonstrates that forced
flags override even direct semantic conflict with the input.

\end{document}